\documentclass[11pt]{article}

\usepackage{color,xspace,wrapfig,fullpage}

\newcommand{\eg}{\emph{e.g.,}\xspace}

\usepackage{multirow,booktabs,amsmath,amsfonts,cite,hyperref,float,epsfig,graphicx,stfloats,url,bm,booktabs,bbding,amssymb,diagbox}
\usepackage[hyphenbreaks]{breakurl}
\usepackage{pifont}
\usepackage{footnote}
\usepackage{verbatim}
\usepackage{subfig}
\usepackage{tikz}
\usepackage{tabularx}
\usepackage{colortbl}
\usepackage{xcolor}
\usepackage{inputenc}

\makesavenoteenv{table}

\begin{document}

\title{\LARGE \bf
Extended vehicle energy dataset (eVED): an enhanced large-scale dataset for deep learning on vehicle trip energy consumption}
	
\author{Shiliang~Zhang\footnote{Chalmers University of Technology, Sweden. Email: shiliang@chalmers.se, \{dyako, fahmi\}@student.chalmers.se} \and Dyako Fatih$^\ast$ \and Fahmi Abdulqadir$^\ast$ \and Tobias Schwarz\footnote{University of Gothenburg, Sweden. Email: gusschwto@student.gu.se} \and Xuehui Ma\footnote{Xi’an University of Technology, China. Email: maxuehuiphd@stu.xaut.edu.cn} }
	
\maketitle

\newcommand{\tabincell}[2]{\begin{tabular}{@{}#1@{}}#2\end{tabular}}

\begin{abstract}
	
This work presents an extended version of the Vehicle Energy Dataset (VED), which is a openly released large-scale dataset for vehicle energy consumption analysis. Compared with its original version, the extended VED (eVED) dataset~\footnote{A description of our eVED dataset can be found at \url{https://github.com/zhangsl2013/eVED}, and users can download our data and code via \textbf{git clone https://Datarepo@bitbucket.org/datarepo/eved\_dataset.git}} is enhanced with accurate vehicle trip GPS coordinates, serving as a basis to associate the VED trip records with external information~\eg road speed limit and intersections, from accessible map services to accumulate attributes that is essential in analyzing vehicle energy consumption. In particularly, we calibrate all the GPS trace records in the original VED data, upon which we associated the VED data with attributes extracted from the Geographic Information System (QGIS), the Overpass API, the Open Street Map API, and Google Maps API. The associated attributes include 12,609,170 records of road elevation, 12,203,044 of speed limit, 12,281,719 of speed limit with direction (in case the road is bi-directional), 584,551 of intersections, 429,638 of bus stop, 312,196 of crossings, 195,856 of traffic signals, 29,397 of stop signs, 5,848 of turning loops, 4,053 of railway crossings (level crossing), 3,554 of turning circles, and 2,938 of motorway junctions. With the accurate GPS coordinates and enriched features of the vehicle trip record, the obtained eVED dataset can provide a precise and abundant medium to feed a learning engine, especially a deep learning engine that is more demanding on data sufficiency and richness. Moreover, our software work for data calibration and enrichment can be reused to generate further vehicle trip datasets for specific user cases, contributing to deep insights into vehicle behaviors and traffic dynamics analyses. We anticipate that the eVED dataset and our data enrichment software can serve the academic and industrial automotive section as apparatus in developing future technologies.

\end{abstract}

\section{Introduction}

The transportation section is one of the greatest energy consumer, and the investigations of on-road vehicle behavior and traffic dynamics can lead to deep insight into energy-efficient route planning. However, there is an insufficiency of open large-scale dataset dedicated to vehicle energy consumption research. Particularly, there lacks a large dataset with features of significant richness of attributes, which fits with deep learning purpose aiming to dig the data deep and extract useful patterns from it. Though various vehicle datasets have been used in the research community~\cite{CALEARO2021111518}, not all of them are openly released that can be leveraged by a wide range of studies. Table~\ref{table: dataset list} lists some publicly available datasets for vehicle energy and mobility analysis, with their scale and feature amount compared.

\begin{table}[htp]
	\centering
	\caption{\label{table: dataset list}A list of related open-source vehicle trip datasets}
	\renewcommand\arraystretch{1.6}
	\resizebox{0.95\textwidth}{!}{
		\begin{tabular}{|l|l|l|l|l|}
			\cline{1-5}
			Dataset name with citations of use in research community & \tabincell{l}{Amount of\\ records} & \tabincell{l}{Amount of\\features} & \tabincell{l}{Recorded from\\ diverse data\\ owners or not} & \tabincell{l}{Contains energy\\ consumption related\\ features or not} \\
			\cline{1-5}
			\href{https://ieee-dataport.org/open-access/battery-and-heating-data-real-driving-cycles}{\tabincell{l}{Battery and heating data in real driving cycles}}~\cite{wevj12030115,DBLP:conf/icce-berlin/PauDGS21,NOGAY2022100590} & 900,000+ & 48 &  No & Yes \\
			\cline{1-5}
			\href{https://github.com/gsoh/VED}{\tabincell{l}{Vehicle Energy Dataset }}~\cite{9201359,DBLP:conf/bigdataconf/PrehoferM20,DBLP:conf/ssd/PetkeviciusSCT21,9682536,DBLP:journals/access/ShafeeFMAAA20,DBLP:conf/ijcnn/LinZWX21,DBLP:conf/aina/WerghuiKOK21,doi:10.1061/9780784483565.005,HAN2022115022} & 20,000,000+ & 18 & Yes & Yes \\
			\cline{1-5}
			\href{https://data.mendeley.com/datasets/wykht8y7tg/1}{\tabincell{l}{Panasonic 18650PF Li-ion Battery Data}}~\cite{DBLP:journals/tie/ChemaliKPAE18,7993319} &  1,000,000+ & 10 &  No & No \\
			\cline{1-5}
			\href{https://data.matr.io/1/projects/5c48dd2bc625d700019f3204}{\tabincell{l}{Battery cycle life data}}~\cite{DBLP:journals/corr/abs-2110-09687} & 800,000+ & 6 &  No & No \\
			\cline{1-5}
			\href{https://www.kaggle.com/vmarrapu/ev-battery-level-vs-distance-approximation}{\tabincell{l}{EV battery level vs distance approximation}} & 108 & 4 & No & No \\
			\cline{1-5}
			\href{https://ieee-dataport.org/documents/new-dynamic-stochastic-ev-model-power-system-planning-applications}{\tabincell{l}{A New Dynamic Stochastic EV Model\\ for Power System Planning Applications}} & 364 & 2 &  Yes & No \\
			\cline{1-5}
			\href{https://ieee-dataport.org/documents/automotive-li-ion-cell-usage-data-set\#files}{\tabincell{l}{Automotive Li-ion Cell Usage Data Set}}~\cite{DOSREIS2021100081} & 68,771 & 5 & Yes & Yes \\
			\cline{1-5}
			\href{https://ieee-dataport.org/documents/electric-vehicle-speed-pedal-accel-voltage-and-current-data-over-four-different-roads}{\tabincell{l}{Electric vehicle speed, pedal, acceleration, voltage,\\ and current data over four different roads}}~\cite{electronics9020278} & 163,984 & 6 &  No & Yes \\
			\cline{1-5}
			\href{https://view.officeapps.live.com/op/view.aspx?src=https\%3A\%2F\%2Favt.inl.gov\%2Fsites\%2Fdefault\%2Ffiles\%2Fpdf\%2Fphev\%2F2013ChevroletVoltChargingData.xlsx\&wdOrigin=BROWSELINK}{\tabincell{l}{Vehicle charging system testing for Chevrolet Volt 2013}} & 29,315 & 4 &  No & Yes \\
			\cline{1-5}
			\href{https://view.officeapps.live.com/op/view.aspx?src=https\%3A\%2F\%2Favt.inl.gov\%2Fsites\%2Fdefault\%2Ffiles\%2Fpdf\%2Fphev\%2F2013FordCMaxEnergiChargingData.xlsx\&wdOrigin=BROWSELINK}{\tabincell{l}{Vehicle charging system testing for Ford C-Max Energi 2013}} & 15,352 & 3 &  No & Yes \\
			\cline{1-5}
			\href{https://view.officeapps.live.com/op/view.aspx?src=https\%3A\%2F\%2Favt.inl.gov\%2Fsites\%2Fdefault\%2Ffiles\%2Fpdf\%2Fphev\%2F2013FordFocusChargingData.xlsx\&wdOrigin=BROWSELINK}{\tabincell{l}{Vehicle charging system testing for Ford Focus Electric}} & 3,601 & 4 &  No & Yes \\
			\cline{1-5}
			\href{https://view.officeapps.live.com/op/view.aspx?src=https\%3A\%2F\%2Favt.inl.gov\%2Fsites\%2Fdefault\%2Ffiles\%2Fpdf\%2Fphev\%2F2013FordFusionEnergiChargingData.xlsx\&wdOrigin=BROWSELINK}{\tabincell{l}{Vehicle charging system testing for Ford Fusion Energi}} & 1,663 & 4 &  No & Yes \\
			\cline{1-5}
			\href{https://view.officeapps.live.com/op/view.aspx?src=https\%3A\%2F\%2Favt.inl.gov\%2Fsites\%2Fdefault\%2Ffiles\%2Fpdf\%2Fphev\%2F2013NissanLeafElectricChargingData.xlsx\&wdOrigin=BROWSELINK}{\tabincell{l}{Vehicle charging system testing for Nissan Leaf}} & 3,343 & 6 &  No & Yes \\
			\cline{1-5}
			\href{https://view.officeapps.live.com/op/view.aspx?src=https\%3A\%2F\%2Favt.inl.gov\%2Fsites\%2Fdefault\%2Ffiles\%2Fpdf\%2Fphev\%2F2013ToyotaPriusPlugInChargeData.xlsx\&wdOrigin=BROWSELINK}{\tabincell{l}{Vehicle charging system testing for Toyota Prius Plug-in}} & 1224 & 4 &  No & Yes \\
			\cline{1-5}
		\end{tabular}
	}
\end{table}

Among the accessible databases, the Vehicle Energy Dataset (VED)~\cite{9262035} is the most promising one with its large volume of vehicle driving records and pertinent features toward energy consumption. Nevertheless, the attributes in the VED dataset are records of internal vehicle features like speed, GPS coordinates, engine revolution, vehicle device temperature, etc., without external yet essential information~\eg the speed limit the vehicle is subjected to, the elevation and the road slope, and the traffic lights, which plays an important role in traffic profile and vehicle energy consumption behavior. On the other hand, we notice that GPS records in the VED data are corrupted by noise to different levels, possibly because of a vibrating and moving environment for the on-board diagnostics (OBD) sensors logging the data. Such inaccurate GPS records limit the feasibility to associate the VED dataset with external attributes aforementioned and to extend the VED dataset to an enriched version suitable for deep learning, which is demanding on the data sufficiency and richness. 

To address this concern, we enhance the VED data by generating an extended version of VED abbreviated as eVED. We first conduct map matching to obtain accurate GPS for the VED dataset, using the Hidden Markov Model (HMM) method~\cite{DBLP:conf/gis/NewsonK09} to snap the original coordinates to the most likely road segment represented by a timestamped sequence of latitude-longitude points. Then using the matched GPS, we extract the speed limits corresponding to GPS with a joint software effort on Open Street Map (OSM)~\cite{DBLP:journals/corr/abs-2011-13556}, Quantum Geographic Information System (QGIS)~\cite{DBLP:journals/ijgi/LimaFEAB21}, and Overpass Turbo~\footnote{\url{https://overpass-turbo.eu/}}. The elevation is associated with the records in the VED dataset using the Elevation API~\footnote{\url{https://developers.google.com/maps/documentation/elevation/start}} from Google Maps platform. We also identify intersections in the VED records via a combined querying and processing using QGIS and OSM. The bus stops are also recognized in the vehicle traces consisted in the VED dataset through QGIS operations. Finally, we collect a set of road elements that might cause a change in the vehicle speed, such as stop signs, pedestrian crossing, railway crossing, and traffic lights, and we summarize those road elements as focus points in this work. With the map matching and data enrichment efforts, we provide the eVED dataset as an extended large-scale database with enhanced record accuracy and enriched vehicle trip attributes, which can serve as an accurate apparatus in vehicle energy consumption analysis and empower deep learning engines toward such analysis.

This paper is organized as follows: Section~\ref{section: gps_match} describes the map matching to obtain accurate GPS records for the VED dataset. Based on the matched vehicle GPS, we present the extraction and association of external features of speed limit, elevation, intersection, bus stop, and focus point in section~\ref{section: speed_limits},~\ref{section: elevation},~\ref{section: intersection},~\ref{section: bus_stop}, and~\ref{section: focussing point}, respectively. Section~\ref{section: usecase} demonstrates three use cases based on the eVED dataset, one as a statistical vehicle energy consumption estimation method, another as an anlysis on vehicle speed distribution regarding traffic tension, and the last one as a deep learning example for vehicle speed estimation, exemplifying the potential of our eVED data to support automotive research. We conclude this work in section~\ref{section: conclustion}.

\begin{figure}[!htbp]
	\centering
	\includegraphics[width=0.38\textwidth]{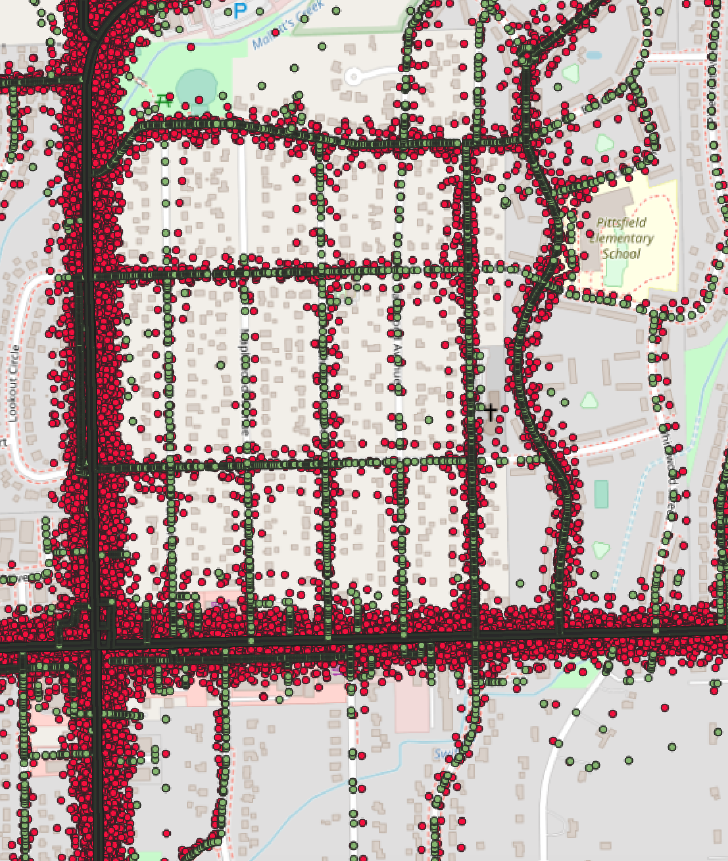}\\
	\caption{\label{fig: gsp matching}The recorded vehicle trace GPS coordinates in the raw VED dataset (in red dots) and their potentially real location estimated by our map matching efforts (in green dots). Notice that quite a number of recorded GPS coordinates fall out a road where they are actually on.}
\end{figure}

\section{Map matching}\label{section: gps_match}

A common problem with vehicle GPS traces is that the recording GPS is prone to disturbance, leading to a recorded position in a GPS trace possibly to be off by a few meters if not more. We show an example of such divergence in the raw VED dataset depicting to what extent the difference between recorded and potential truth GPS coordinates can be in Figure~\ref{fig: gsp matching}. An inaccurate GPS makes it hard to relate a vehicle position with infrastructures affecting vehicle behaviors like intersection, bus stop, pedestrian crossing, etc., thus blocking further analysis on vehicle and traffic. 

\begin{figure}[!htbp]
	\centering
	\includegraphics[width=0.4\textwidth]{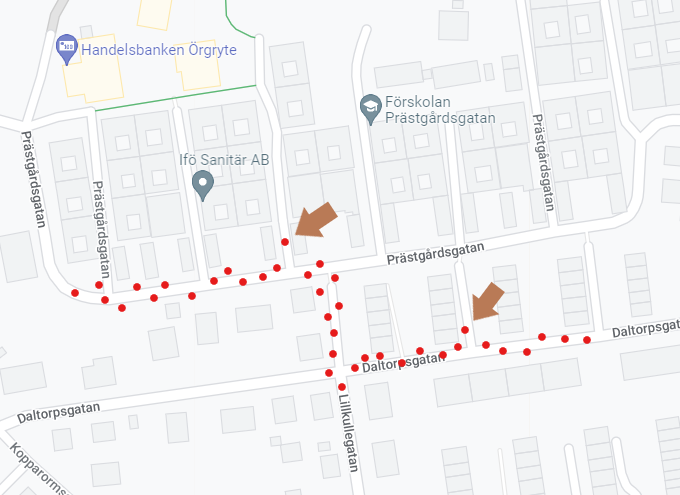}\\
	\caption{\label{fig: mis match example}The inaccuracy of the approach to match GPS coordinates to their nearest road segment, demonstrated by the two pointed-out coordinates which are closer to a road not consistent with the actual moving trace.}
\end{figure}

\begin{figure}[!htbp]
	\centering
	\includegraphics[width=0.4\textwidth]{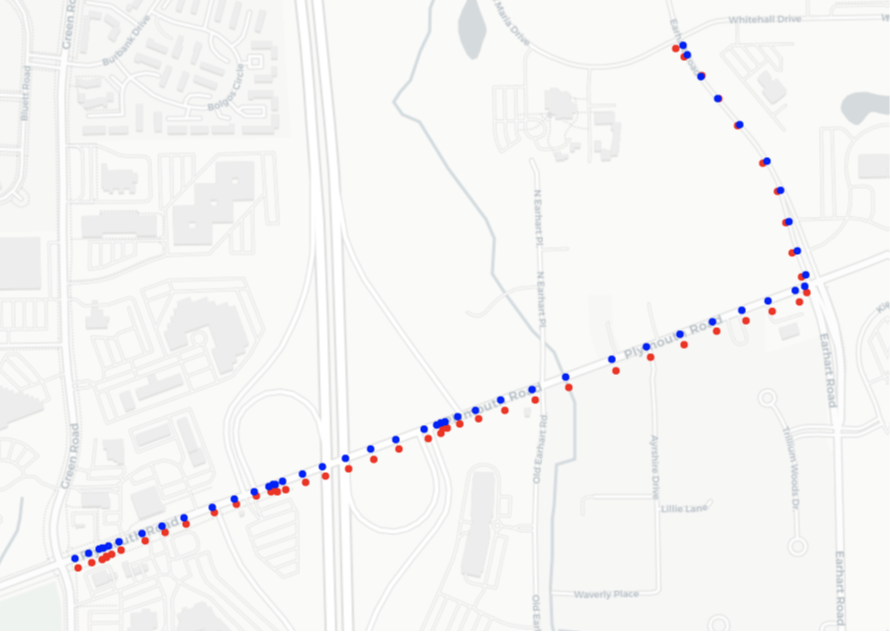}\\
	\caption{\label{fig: gsp snap}The recorded vehicle trace GPS coordinates (in red dots) and their most likely counterpart (in blue dots) based on the HMM approach.}
\end{figure}

A potential solution to calibrate the GPS is to match every GPS coordinate to its nearest road segment. However, such solution might not always be accurate since some recorded coordinates might appear to lie on or closer to another road segment other than the real one, as illustrated in Figure~\ref{fig: mis match example}. This work leverages the Hidden Markov Model (HMM) approach to snap the GPS trace records to the most likely consecutive road segment, which is represented by a time-stamped sequence of latitude-longitude coordinates that are the closest to the recorded GPS coordinates, as shown in Figure~\ref{fig: gsp snap}.

We implement the HMM approach using Valhalla~\footnote{\url{https://github.com/valhalla/valhalla}}, which is an open-source routing engine which can be used along with Open Street Map (OSM) data to match GPS coordinates. Particularly, we use the Meili library in Valhalla that, for a given sequence of GPS points, Meili finds the nearest most-likely sequence of candidates of consecutive road segments call Viterbi path, which is generated based on the Viterbi algorithm~\cite{DBLP:journals/ett/YouLCR16}.

The GPS matching is not always working well, and we mark the matched GPS coordinates in the generated eVED dataset with different labels of fully matched, interpolated, and unmatched. We mark the interpolated GPS coordinates in the generated data file since the raw VED data contains many consecutive duplicate coordinates, in which case we label the first coordinate as matched and interpolated for the rest duplicates, although they are matched with the same point. Those matched coordinates locating out of any road is marked as unmatched in the eVED dataset. Overall, the coordinate matching manages to snap over 99\% of the GPS records in the raw VED dataset to coordinates on the road.

\section{Speed limit}\label{section: speed_limits}

This section associates the raw VED records with speed limit of the road where the vehicle is driving on, based on the calibrated GPS coordinates obtained in section~\ref{section: gps_match}. To facilitate the extraction of speed limit information, we first download the drivable road data using Quantum Geographic Information System (QGIS), and then we add the line layer from Overpass Turbo containing the road speed limit information to the QGIS canvas, and finally append the speed limit to the raw VED dataset by the QGIS software.

The speed limit of a road in the Overpass line layer is defined by the tag ``maxspeed", in addition to which there are other tags indicating different classes of speed limit, such as advisory and practical. Moreover, there are scenarios where a single road segment has different speed limit depending on the traveling direction. To wrap up those information in the eVED dataset, we add a column in the dataset defined as ``class" for the speed limit we extracted, detailed in Table~\ref{table: speed limit classes}. We also add a column as ``speed limits with direction" to include only the speed limit of the direction the car was driving along, which is recognized via the calculation of the initial bearing $\theta$ in equation~\ref{equation: theta}~\cite{miller2017implementation}, where the bearing is calculated for the road segment as $\theta_{1}$, and for the coordinates of the trip along that road segment as $\theta_{2}$. The direction of the travel is forward if $cos(\theta_{1}-\theta_{2})>0$, otherwise backward.

\begin{equation}\label{equation: theta}
		\begin{split}
		 		\theta=atan2(sin\Delta\lambda\cdot cos\phi_{2},\;\; cos\phi_{1}\cdot sin\phi_{2}-sin\phi_{1}\cdot cos\phi_{2}\cdot cos\Delta\lambda)\\
		\end{split}
\end{equation}
where $\phi_{1}, \lambda_{1}$ and $\phi_{2}, \lambda_{2}$ are the starting and ending latitude-longitude, respectively, and $\Delta=\lambda_{2}-\lambda_{1}$.

\begin{table}[htp]
		\centering
		\caption{\label{table: speed limit classes}Description of speed limit classes. The tag ``maxspeed:forward" means that the direction of travel is the same as that of the ordered sequence of points in the LineString of that road segment, whereas ``maxspeed:backward" means the opposite. The speed limit class labeled as 1 represents the case where there are no explicit speed limit signs, therefore the default speed limits apply for the city of Michigan where the VED data were recorded as follows: motorway 70 mph, trunk 55 mph, primary 55 mph, secondary 45 mph, tertiary 35 mph, unclassified 55 mph, residential 25 mph, and service 25 mph.}
		\resizebox{0.9\textwidth}{!}{
				\begin{tabular}{|l|l|l|l|}
						 \hline
						 Class & \tabincell{l}{OSM tag from Overpass\\ line layer} & Description & \tabincell{l}{Proportion of raw VED\\ records (\%)} \\
						 \hline
						 -1 & \tabincell{l}{maxspeed:forward\\maxspeed:backward} & \tabincell{l}{maximum legal speed limit depending\\on the direction of travel} & 0.569\\
						 \hline
						 0 & maxspeed & maximum legal speed limit & 81.7 \\
						 \hline
						 1 & - & default speed limits & 14.4 \\
						 \hline
						 2 & maxspeed:advisory & \tabincell{l}{posted advisory speed limit which is\\not legally binding} & 0.929 \\
						 \hline
						 3 & maxspeed:practical & \tabincell{l}{realistic average speed estimate\\when there are no legal speed limits} & 0.027 \\
						 \hline
					\end{tabular}
			}
\end{table}

\section{Elevation}\label{section: elevation}

The integrating of elevation information in the eVED dataset is intuitive since it has a direct impact on vehicle behavior, e.g., traversing up an incline requires more energy than a flat surface or decline. To associate the raw VED data records with elevation, we leverage a service named Elevation API from Google Maps platform which can provide the elevation information for a given GPS coordinate in the raw VED data.

The raw VED dataset contains about 22 million rows of data record, each of which is made up of several attribute columns whose values are gathered at different intervals. Due to such different sampling intervals, part of the recorded GPS coordinates are duplicated across several consecutive records because of the lower polling rate of GPS. Furthermore, duplication of coordinates induces the extracted elevation data to be stepped. To mitigate such impact, we smooth the extracted elevation using a ``five-point" average approach, i.e., we take the average of five points, which consist of two points preceding and two succeeding including the point under consideration, as the smoothed elevation of the GPS point in question. Note that we keep both the raw and the smoothed elevation in the eVED dataset files for reference.

With the smoothed elevation information for vehicle traces, we calculate the gradients based on two consecutive vehicle GPS points in a procedure shown in Table~\ref{table: procedure for elevation gradient}.

\begin{table}[htp]
	\centering
	\caption{\label{table: procedure for elevation gradient}Procedures for the calculation of road gradient based on elevation.}
\begin{tabular}{p{0.96\textwidth}}
	\toprule
	\textbf{Constant:} $N$ as the total number of records in the raw VED dataset\\
	\\
	\textbf{for} $i\in\{1,2,3,...,N-1\}$\\
	\qquad take the $i$-th and its succeeding $(i+1)$-th coordinate\\
	\qquad query the elevation h($i$) and h($i+1$) for the $i$-th and the $(i+1)$-th coordinate, respectively\\
	\qquad \textbf{if} h($i$) = h($i+1$):\\
	\qquad \qquad set the road gradient g($i$) for the $i$-th coordinate as 0;\\
	\qquad \textbf{else}\\
	\qquad \qquad calculate the geodesic distance $D_{g}$ between the  $i$-th and $(i+1)$-th coordinate using \\
	\qquad \qquad the GeoPy's distance function~\footnote{\url{https://geopy.readthedocs.io/en/stable/\#module-geopy.distance}}, which follows Karney's approach~\cite{karney2013algorithms}.\\
	\qquad \qquad calculate the road gradient g($i$) as:\\
	\qquad \qquad g($i$)=(h($i+1$)-h($i$))/$D_{g}$\\
	\qquad \textbf{end}\\
	\textbf{end}\\
	\bottomrule
\end{tabular}
\end{table}

\section{Intersection}\label{section: intersection}

Based on the calibrated GPS records, we associate the raw VED data records with indicator showing whether a vehicle is driving through an intersection, and mark the record in question when the answer is yes. The extraction of intersection is conducted using the Quantum Geographic Information System (QGIS), which enable the recognition of street intersections from Open Street Map (OSM) and consequently label the GPS coordinates of all intersections for a given vehicle mobility trace. 

\begin{figure}[!htbp]
	\centering
	\includegraphics[width=0.35\textwidth]{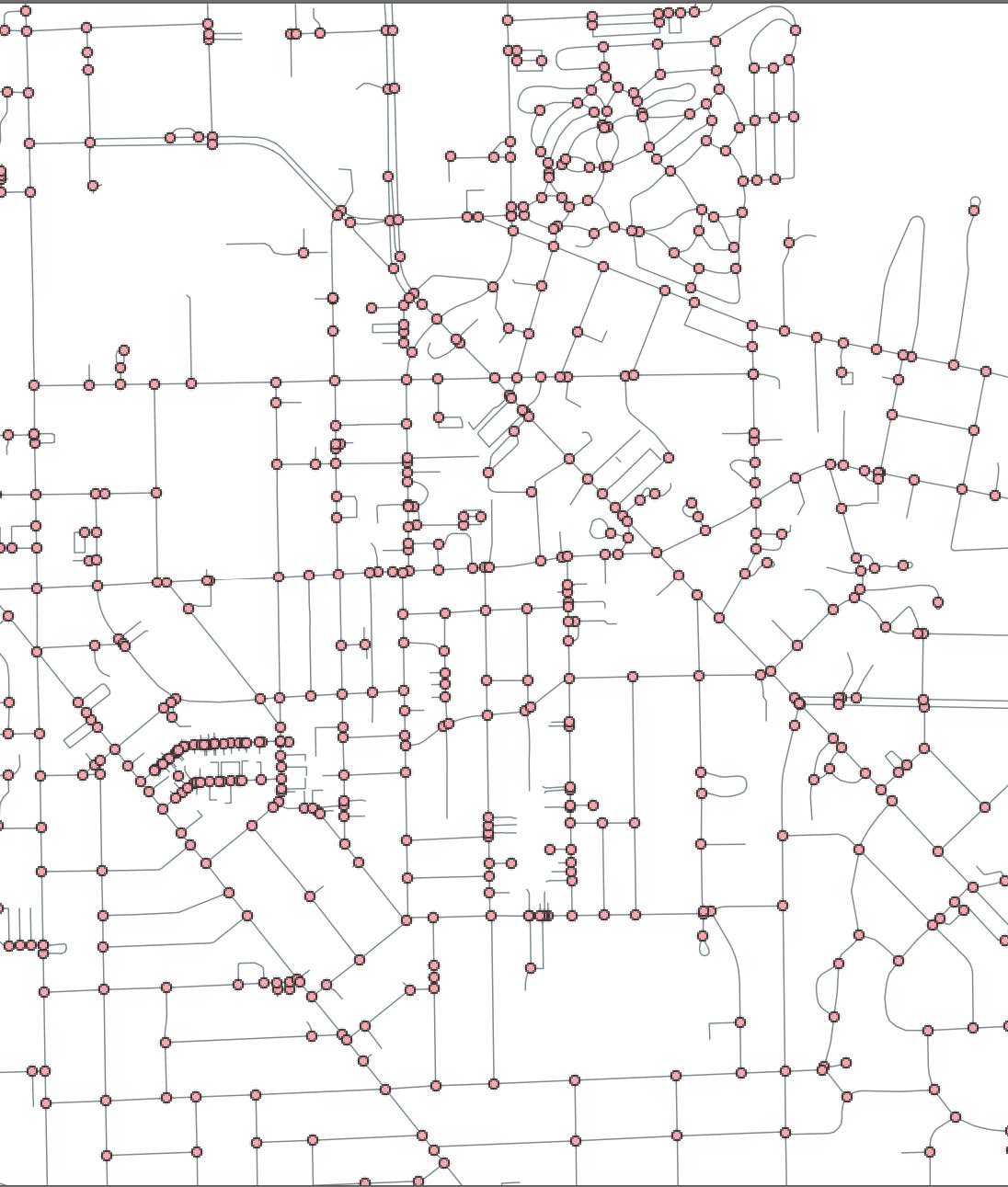}\\
	\caption{\label{fig: intersection}An example of intersections extracted from Open Street Map for the roads in Ann Arbor, Michigan, USA.}
\end{figure}

We first generate a layer using QGIS that covers all the area that the vehicle data is recorded, whereby we locate all the intersection coordinates on the layer. In this case, such coordinates constitute the universal set of intersections in the city of Ann Arbor as shown in Figure~\ref{fig: intersection}. Using this universal set, we can traverse all the vehicle data records and mark a record as an intersection according to the record's GPS coordinate. 

\section{Bus stops}\label{section: bus_stop}

The bus stop locations in a specific area can be downloaded from the Open Street Map (OSM) as a point layer, based on which we use QGIS to extract the coordinates of those locations. Before the extraction, we unify the coordinates of both the original and the OSM layer data into the EPSG:4326 coordinate reference system (CRS)~\cite{HARE201836}, which allows us to use meters as the unit of measurement during the calculation. 

\begin{figure}[!htbp]
	\centering
	\includegraphics[width=0.55\textwidth]{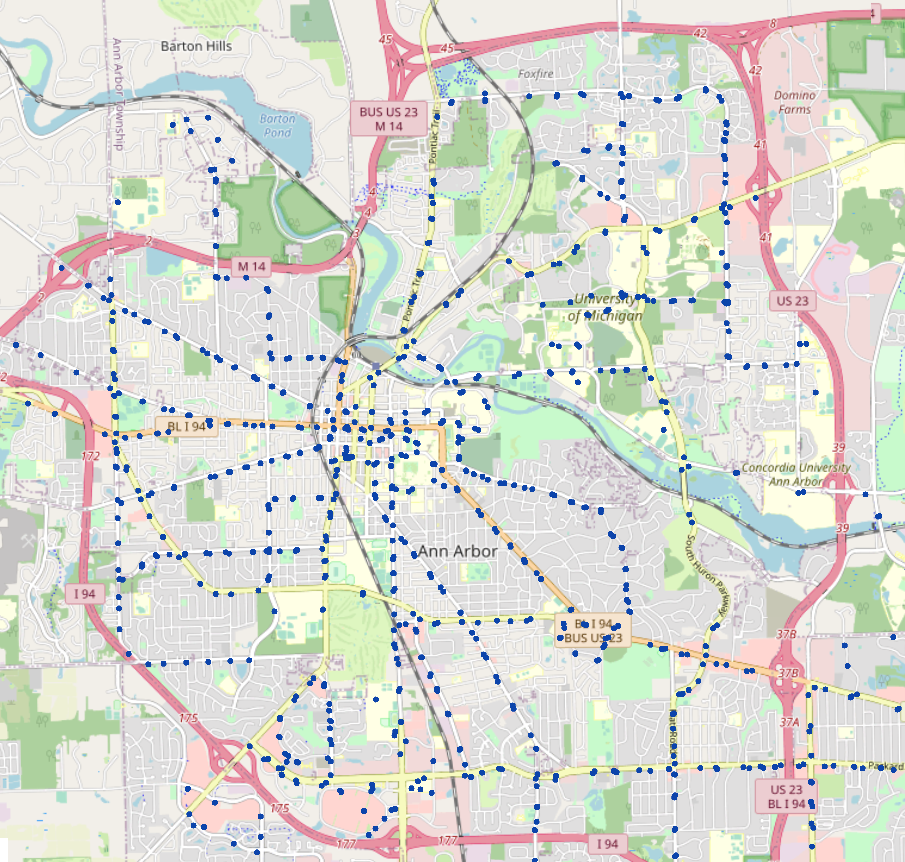}\\
	\caption{\label{fig: busstop}Visualization of the data for bus stop locations in Ann Arbor, Michigan, USA, where each solid blue dot represents a bus stop.}
\end{figure}

The data for bus stops is download from the data catalog of the website of City of Ann Arbor Michigan~\footnote{\url{https://www.a2gov.org/services/data/Pages/default.aspx}}, as shown in Figure~\ref{fig: busstop}. Using the bus stop locations, we create a buffer of 10 meters in radius using QGIS and the coordinates in the raw VED data that are at most 10 meters away from any bus stop are marked, and we generate a single column to include such indications for bus stops in our data files. In total, we traverse the all the vehicle records and recognize 429,638 of them as driving through an intersection.

\section{Focus points}\label{section: focussing point}

The term of focus points in this work encapsulates the location of interest on a road such as traffic signal, pedestrian crossing, stop signs that might cause a change in vehicle speed. Particularly, we consider 16 type of focus points as 1) crossing, 2) traffic signal, 3) stop signs, 4) turning loop, 5) railway crossing, 6) turning circle, 7) motorway junction, 8) hump, 9) lift gate, 10) gate, 11) give way, 12) bollard, 13) level crossing, 14) roundabout 15) mini roundabout, 16) swing gate. Note that all the terms used for focus points are consistent with the definition on the OpenStreetMap Wiki~\footnote{\url{https://wiki.openstreetmap.org/wiki/Main_Page}}.

We first download the point layer containing all the types of focus points considered from OpenStreetMap, and using QGIS we are able to create three-meter radius centered at each focus point. Then a spatial join is conducted between the VED coordinates and those radii we generated, which entails the recognition of vehicle GPS as experiencing a focus point at most three meters away from it. Table~\ref{table: focus points} shows what types and how many focus points are recognized in our eVED dataset.

\begin{table}[htp]
\centering
\caption{\label{table: focus points}Focus points recognized in the eVED dataset.}
\resizebox{0.99\textwidth}{!}{
\begin{tabular}{|c|c|c|c|c|c|c|c|c|c|c|c|c|c|c|c|c|}
\hline
\rotatebox{90}{\textbf{focus point type}} &  \rotatebox{90}{\textbf{bollard}} & \rotatebox{90}{\textbf{bump}} & \rotatebox{90}{\textbf{crossing}} & \rotatebox{90}{\textbf{gate}} & \rotatebox{90}{\textbf{lift gate}} & \rotatebox{90}{\textbf{give way sign}} & \rotatebox{90}{\textbf{hump}} & \rotatebox{90}{\textbf{level crossing}} & \rotatebox{90}{\textbf{mini roundabout}} & \rotatebox{90}{\textbf{motorway junction}} & \rotatebox{90}{\textbf{roundabout}} & \rotatebox{90}{\textbf{stop sign}} & \rotatebox{90}{\textbf{swing gate}} & \rotatebox{90}{\textbf{traffic signal}} & \rotatebox{90}{\textbf{turning circle}} & \rotatebox{90}{\textbf{turning loop}}  \\
\hline
\tabincell{l}{number}  & 1 & 108 & 312,196 & 1,105 & 96  & 985 & 223 & 4,053 & 195 & 2,938 & 81 & 29,397 & 4 & 195,856 & 3,554 & 5,848 \\
\hline
\end{tabular}
}
\end{table}

\section{Three use cases on vehicle behavior analysis using the eVED dataset}\label{section: usecase}

This section showcases three research/analysis scenarios where our eVED dataset can be applied to, that are (i) an approach to estimate vehicle trip energy consumption that can help a energy-efficient route planning and (ii) vehicle speed profiling for traffic tension analysis, and (iii) a deep learning demonstration for estimating vehicle speed range with only public information. We use those examples to show the opportunities that the eVED data can bring to automotive research.

\subsection{Statistical energy consumption estimation}\label{section: usecase1}

\begin{figure}[!htbp]
	\centering
	\includegraphics[width=0.8\textwidth]{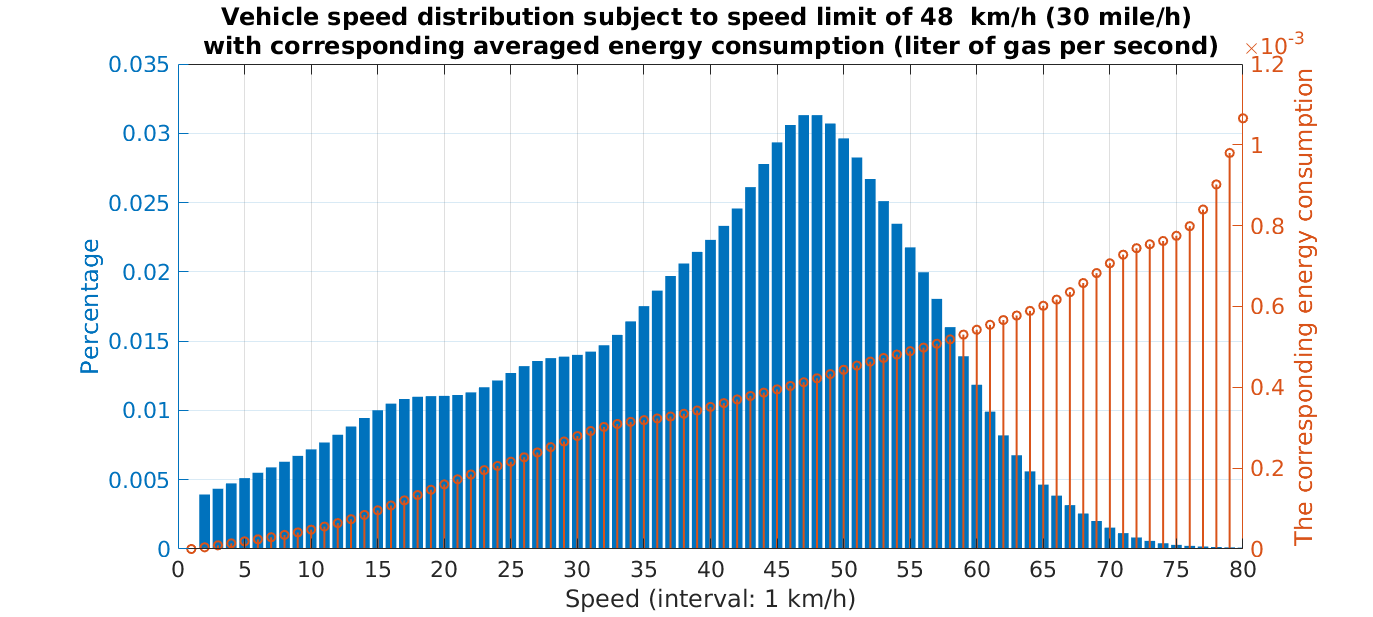}\\
	\caption{\label{fig: speed energy relationship}The speed \& energy distribution subject to the speed limit of 30 mile/h (48 km/h). The blue bars represent the probabilistic density for the vehicle speed subject to the designated speed limit, and the orange stems indicate the average energy consumption for one second's driving under a specific speed.}
\end{figure}

This subsection demonstrates an approach to estimate vehicle trip energy consumption via data statistics of the eVED. Our hypothesis for energy here is the energy consumption is only determined by the vehicle speed. With the simplified hypothesis, we can assign the speed with its corresponding energy consumption and calculate the energy consumption of a trip via the trip's speed profile.

The method is to generate the statistical energy consumption within a certain duration of driving for different speed limits. In this case, we obtain the energy within one second's driving, and add up to a total trip energy consumption within its travel time. The statistical energy consumption is calculated through the distribution of speed subject to a specific speed limit, as an example shown in Figure~\ref{fig: speed energy relationship} for the energy statistics subject to the speed limit of 48 km/h. From the speed \& energy distribution statistics, we can obtain a weighted average energy consumption value for one second's driving under the road with speed limit of 48 km/h, and we can gain such energy consumption values for the other speed limits in a similar way.

\begin{figure}[!htbp]
	\centering
	\includegraphics[width=0.55\textwidth]{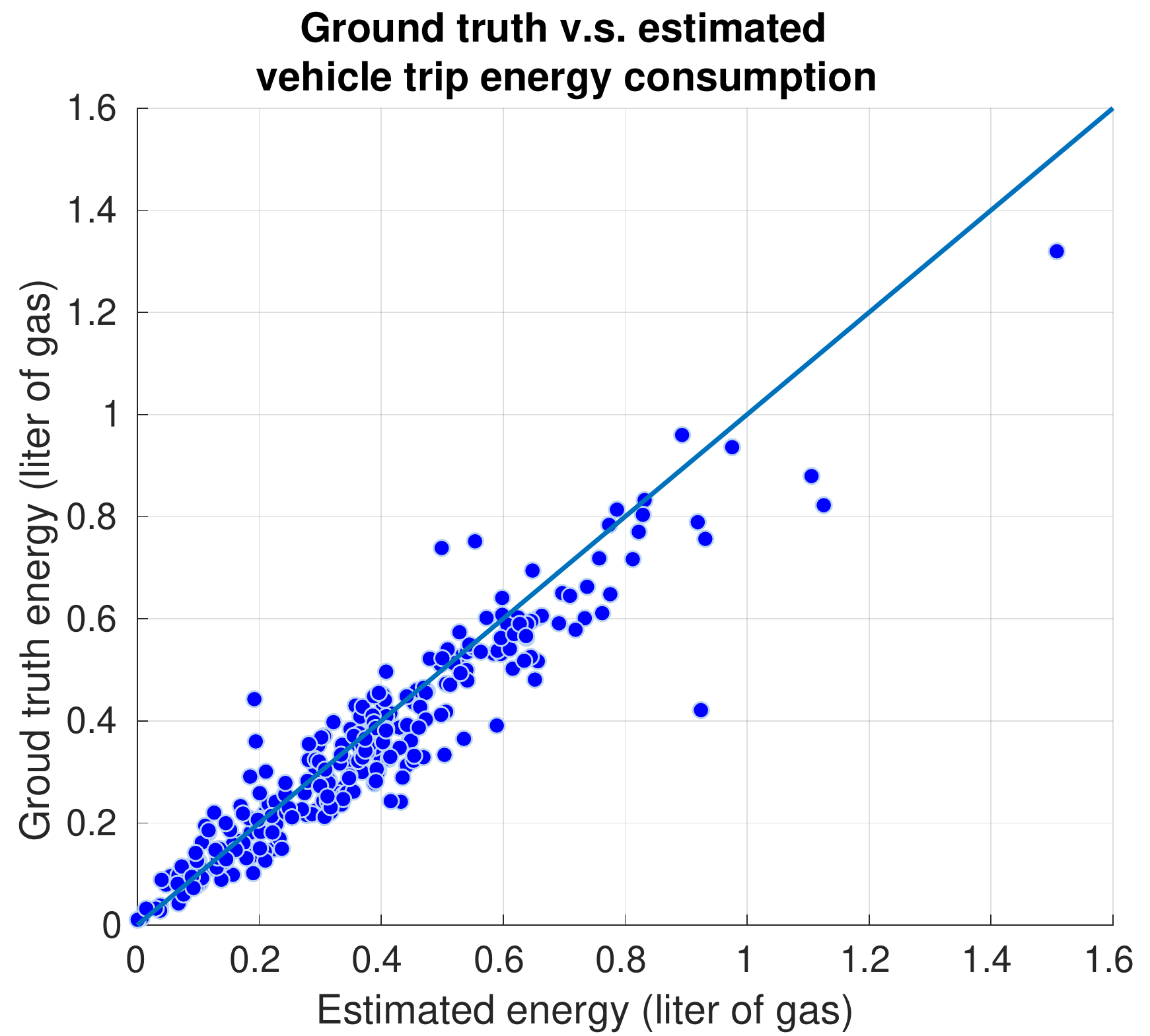}\\
	\caption{\label{fig: estimated vs actual energy}Comparison between the actual and estimated vehicle trip energy consumption.}
\end{figure}

With the statistics of energy consumption for different speed limits obtained, we can estimate the energy consumption for a given vehicle trip by looking at how much time the vehicle drives under different speed limits. To validate this approach, we use 1207 trips of different length from 35 vehicles in generating the energy consumption statistics, and use 277 trips of different length from 7 another vehicles to test the effectiveness of the energy estimation approach. As shown in Figure~\ref{fig: estimated vs actual energy}, the test result manifest a sound estimation accuracy, with a root mean squared error (RMSE) of 0.0781 for the energy estimation of the whole test trips. 

\subsection{Speed distribution analysis regarding traffic tension}\label{section: usecase2}

Using the eVED dataset, this subsection provides an analysis on vehicle speed distribution under a given speed limit. We show an example of vehicle speed distribution subject to the speed limit of 55 mile/h (89 km/h), and elicit insight into traffic tension for the considered speed limit.

\begin{figure}[!htbp]
	\centering
	\includegraphics[width=0.85\textwidth]{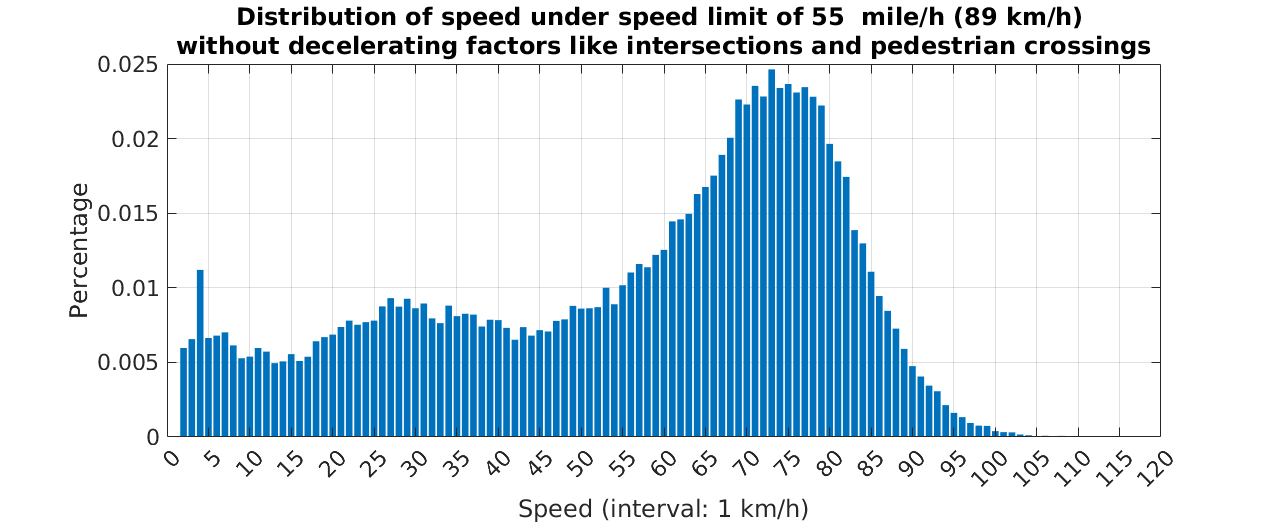}\\
	\caption{\label{fig: free flow distribution}The free flow speed distribution of the vehicle driving under road speed limit of 55 mile/h (89 km/h). Note that the sample used to generate this histogram is free from any factors that can cause a deceleration of the vehicle, like traffic signal, roundabout, crossings, etc.}
\end{figure}

We extract all the speed samples subject to 89 km/h in the eVED dataset and generate a histogram shown in Figure~\ref{fig: free flow distribution}. The histogram shows that the resulted distribution is far from a normal distribution with a bell shape centered around the speed limit of 89 km/h, even though we have removed the speed samples subject to decelerating elements, e.g., traffic signals, before generating the histogram. Particularly, there are about 25\% samples with speed below 50 km/h.

\begin{figure}[!htbp]
	\centering
	\includegraphics[width=0.8\textwidth]{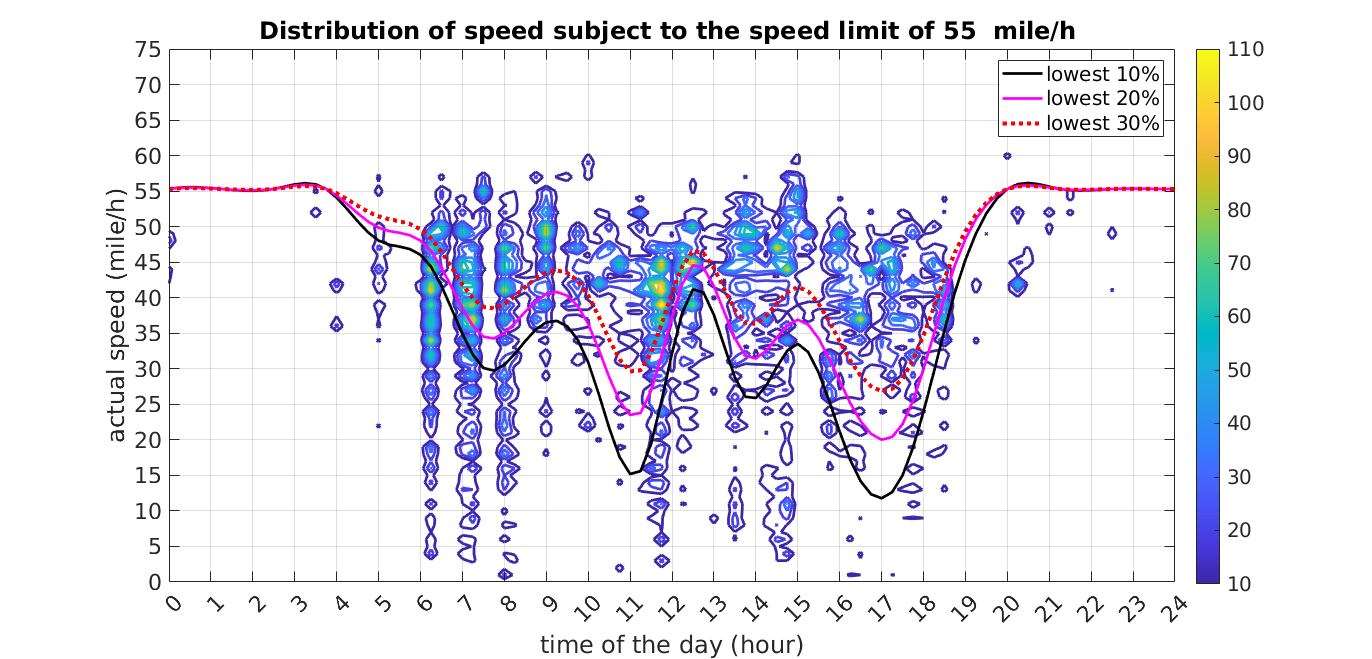}\\
	\caption{\label{fig: contention}Heat map of speed distribution for vehicles driving under road speed limit of 55 mile/h over the time of the day. The three lines in the heat map indicate the lowest 10\%/20\%/30\% population of the speed samples at a specific time of the day.}
\end{figure}

A possible hypothesis is that those speed samples are subject to traffic contention or even jams, as commonly exist in the morning rush hours on the working days. To validate such assumption, we generate the heat map of how the speed population distributes over the time of the day, as shown in Figure~\ref{fig: contention}. The heat map demonstrates several clear time zones where the speeds are much lower than the speed limit appear, e.g., the time from six to nine o'clock in the morning, around twelve in the noon, and around three and six o'clock in the afternoon. Such time zones of lower speed is consistent with daily life expectation where the traffic might be tense during the morning/afternoon when people are heading for working/going back home, and during the noon when people temporally leave their working place and go out for launch.

\subsection{Deep learning for speed estimation}\label{section: usecase3}

This subsection shows how to estimate vehicle speed via training a deep learning model based on the eVED dataset. The idea is to learn a model to estimate the speed for vehicle trip segments, thus we can predict whether there is traffic tension or free traffic flow for a specific route and contribute to the travel time estimation. The training of the model will only use the publicly available information to limit the risk of vehicular privacy-sensitive information leakage, and we list all the input and output features involved in the model training in Table~\ref{table: input-output features}. For the output of the deep learning model, we use the median value of the speed records during two minutes' driving.

\begin{table}[htp]
	\centering
	\caption{\label{table: input-output features}Input and output features description}
	\resizebox{0.65\textwidth}{!}{
		\begin{tabular}{|l|l|l|}
			\cline{2-3}
			\multicolumn{1}{c|}{} & \multicolumn{2}{c|}{features}\\
			\hline
			\multirow{2}{*}{input} & \multicolumn{2}{l|}{\tabincell{l}{time of the day (h); speed limit (km/h); road gradient;\\ whether experiencing an intersection (0 or 1);\\ whether experiencing a bus stop (0 or 1);\\ whether experiencing any focus point shown in\\ Table~\ref{table: focus points} (0 or 1); whether approaching any intersection,\\ bus stop, or focus point within 30 seconds' drive (0 or 1);\\ whether departing any intersection, bus stop, or focus\\ point within 30 seconds' drive (0 or 1)}} \\
			\hline
			output & \multicolumn{2}{l|}{\tabincell{l}{median of the speed in a consecutive 4-minute's trip segment}}\\
			\hline
		\end{tabular}
	}
\end{table}

\begin{figure}[!htbp]
	\centering
	\includegraphics[width=0.7\textwidth]{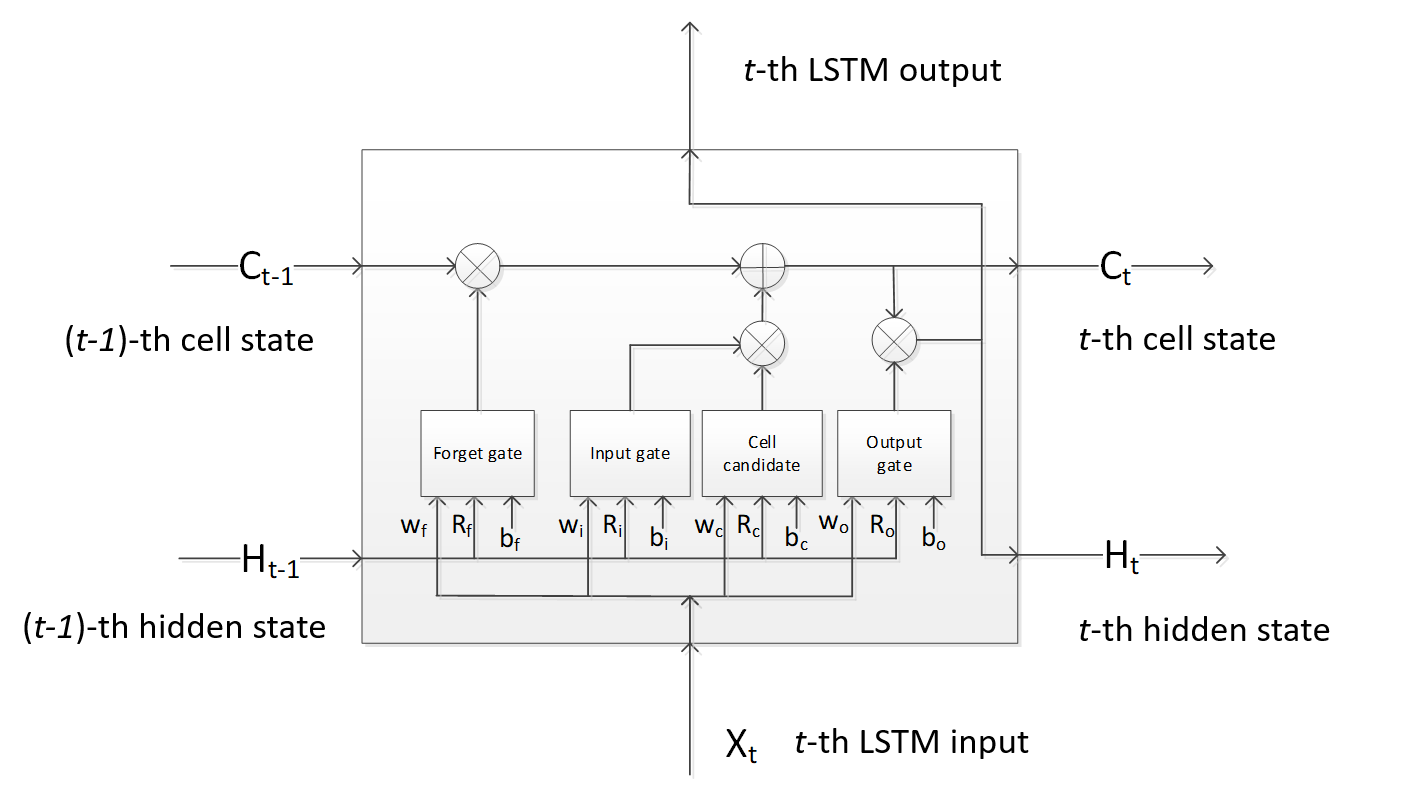}\\
	\caption{\label{fig: lstm_cell}The LSTM layer structure with internal LSTM cell components.}
\end{figure}

We use the long-short-term-memory (LSTM) networks~\cite{DBLP:journals/cee/YangYW22} as the structure of the estimation model as shown in Figure~\ref{fig: lstm_cell}. With the input and output features settled, we can configure the LSTM model accordingly and we initialize the model specifications as shown in Table~\ref{table: lstm specifications}. To facilitate the model training, we extract from the eVED dataset 7,267 trip segments from 308 vehicles, and 951 segments from the 38 vehicles as training and testing dataset, respectively. Each segment consists of 240 sequentially consecutive records of the features listed in Table~\ref{table: input-output features}.

\begin{table}[htp]
	\centering
	\caption{\label{table: lstm specifications}The LSTM model specifications}
	\resizebox{0.7\textwidth}{!}{
		\begin{tabular}{|l|l|}
			\hline
			configuration name & value or dimension\\
			\hline
			LSTM layer number & 1 \\
			\hline
			hidden layer number & 1 \\
			 \hline
			number of hidden units & 5,000 \\
			 \hline
			input features & 8 \\
			 \hline
			state activation function & tanh: $(e^{2x}-1)/(e^{2x}+1)$ \\
			 \hline
			gate activation function for all gates & Sigmoid: $1/(e^{-x}+1)$ \\
			 \hline
			input weight dimension (wi,wf,wc,wo) & 40,000$\times$8 \\
			 \hline
			recurrent weight dimension (Ri,Rf,Rc,Ro) & 40,000$\times$5,000 \\
			 \hline
			bias dimension (bi,bf,bc,bo) & 40,000$\times$1 \\
			 \hline
			fully connected layer number following the LSTM layer & 9 \\
			 \hline
		\end{tabular}
	}
\end{table}

\begin{figure}[!htbp]
	\centering
	\includegraphics[width=0.5\textwidth]{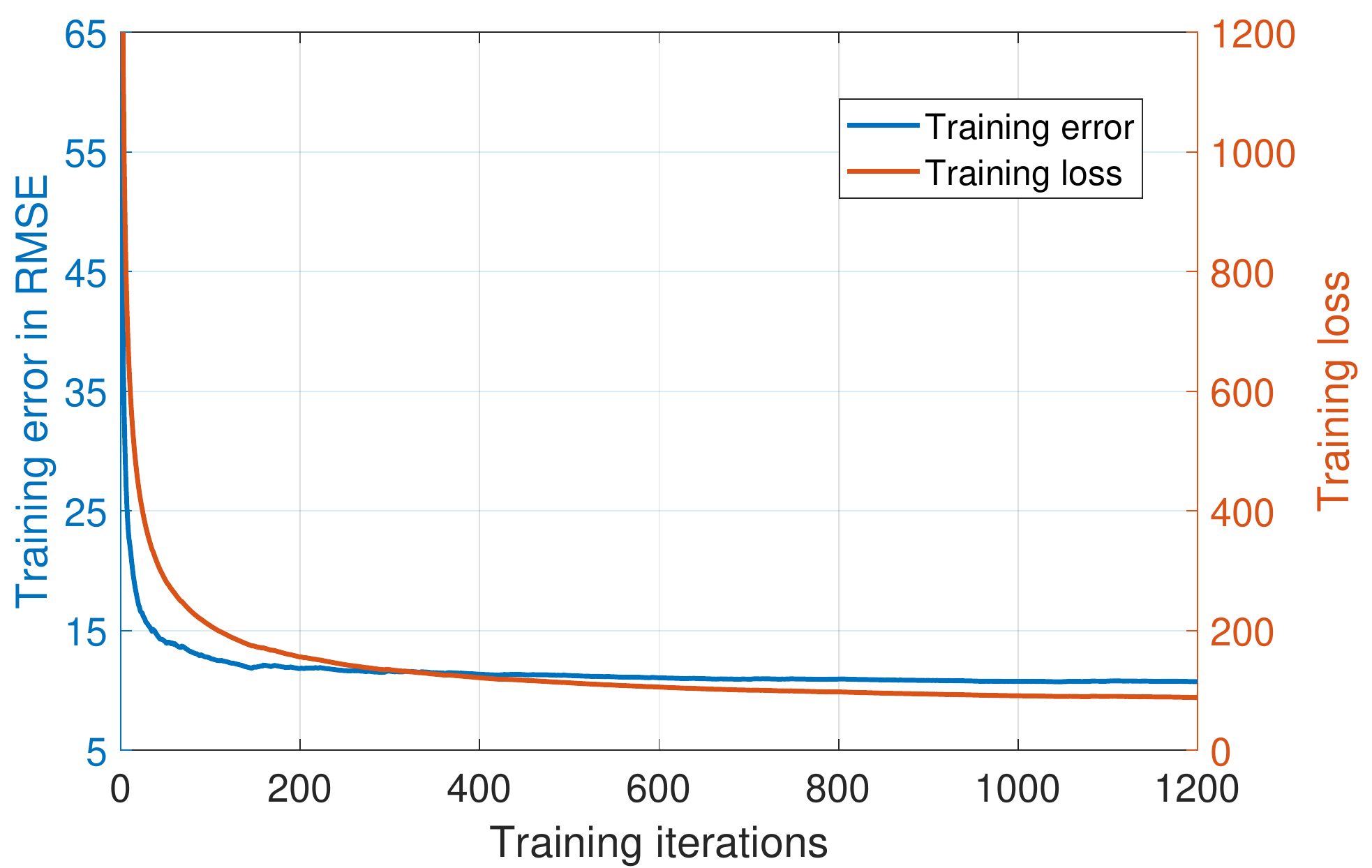}\\
	\caption{\label{fig: lstm training result}Training performance of the trained model in term of training error and training loss.}
\end{figure}

\begin{figure}[!htbp]
	\centering
	\includegraphics[width=0.4\textwidth]{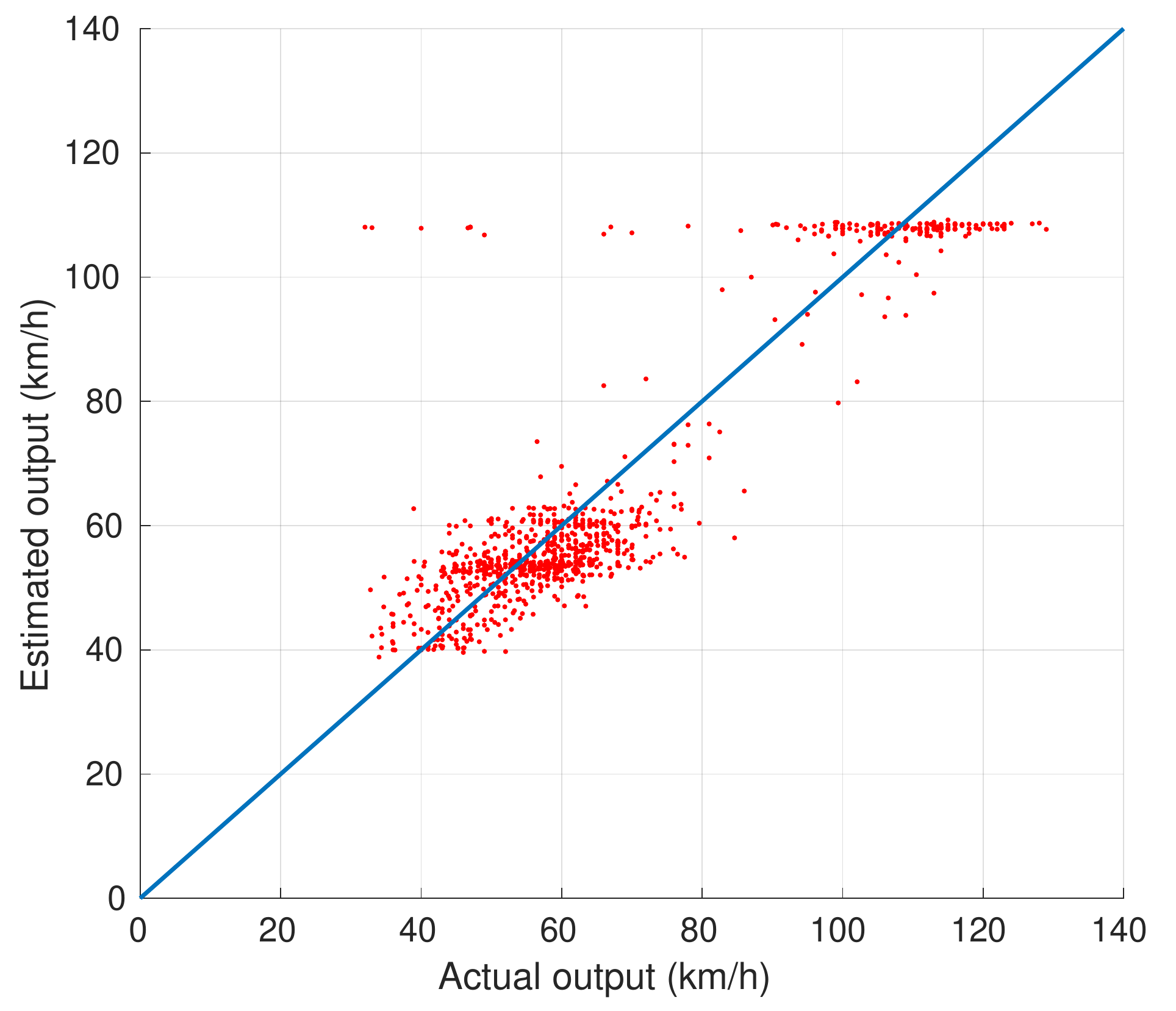}\\
	\caption{\label{fig: lstm test result}Testing performance of the trained model. Note that there are several trip segments with the median speed incorrectly estimated, as shown in the upper-right of this figure. While the majority of the estimated median speeds are consistent with the actual ones with low divergence.}
\end{figure}

We implement the LSTM model using MATLAB 2021b with access to NVidia Tesla A100 GPUs, and we show the training performance in Figure~\ref{fig: lstm training result} in regard with training loss and the root mean squared error (RMSE) of the prediction. Figure~\ref{fig: lstm test result} provides the prediction performance of the trained model on the testing dataset, manifesting that using only publicly available information of vehicle trip data, the estimation of speed for trip segments via a deep learning model can be useful with prediction RMSE of 8.9078 for the test dataset.

\section{Conclusions}\label{section: conclustion}

We present an eVED dataset which is an enhanced and enriched version of the Vehicle Energy Dataset (VED), where we provide precise GPS records based on which we further associate the raw VED data with various features that are essential to vehicle behavior and traffic dynamics. As a large-scale dataset with diversities of features both from internal and external of vehicles, the eVED dataset is potential to serve varieties of vehicle behavior research, particularly for deep learning purposes where the learning engine requires high abundance and richness of the data. Moreover, our software generating the external features for the eVED, which we openly released, can be reused in consumer cases to accumulate tailored datasets to accomplish their analysis. We demonstrate the usage of the eVED dataset via an estimation of vehicle trip energy consumption, an analysis on vehicle speed distribution regarding traffic tension, and a deep learning model to estimate moving vehicle speed only using public information to limit the risk of privacy disclosure. Beyond our demonstrations, we anticipate the eVED dataset to support more advanced vehicle and traffic research and contribute further insights into an efficient transportation network.

\section*{Acknowledgement}

Shiliang Zhang's work was supported by the project `Privacy-Protected Machine Learning for Transport Systems' of Area of Advance Transport and Chalmers AI Research Centre (CHAIR). The computations were enabled by resources provided by the Swedish National Infrastructure for Computing (SNIC) at C3SE partially funded by the Swedish Research Council through grant agreement no. 2018-05973. We also thank Elad Schiller for discussions that help improve the presentation.

\bibliographystyle{IEEEtran}
\bibliography{bibfile}

\end{document}